\documentclass[lettersize,journal]{IEEEtran}
\usepackage{amsmath,amsfonts}
\usepackage{algorithmic}
\usepackage{algorithm}
\usepackage{array}
\usepackage[caption=false,font=normalsize,labelfont=sf,textfont=sf]{subfig}
\usepackage{textcomp}
\usepackage{stfloats}
\usepackage{url}
\usepackage{bbding}
\usepackage{verbatim}
\usepackage{graphicx}
\usepackage{amssymb}
\usepackage{cite}
\usepackage{booktabs}
\begin{document}
	
	\title{Graph Information Bottleneck for Remote Sensing Segmentation}
	
\author{Yuntao~Shou, Wei~Ai, 
	Tao~Meng, Nan Yin
	\thanks{Corresponding Author: Tao Meng~(mengtao@hnan.edu.cn)}
	\IEEEcompsocitemizethanks{\IEEEcompsocthanksitem Y. Shou, W. Ai,~ and~T. Meng are with School of computer and Information Engineering, Central South University of Forestry and Technology, Hunan 410004,
		China. (shouyuntao@stu.xjtu.edu.cn,~aiwei@hnu.edu.cn, mengtao@hnu.edu.cn)
		\IEEEcompsocthanksitem N. Yin is with Mohamed bin Zayed University of Artificial Intelligence, UAE. (nan.yin@mbzuai.ac.ae)
		}
}

	\maketitle
	
\begin{abstract}
Remote sensing segmentation has a wide range of applications in environmental protection, and urban change detection, etc. Despite the success of deep learning-based remote sensing segmentation methods (e.g., CNN and Transformer), they are not flexible enough to model irregular objects. In addition, existing graph contrastive learning methods usually adopt the way of maximizing mutual information to keep the node representations consistent between different graph views, which may cause the model to learn task-independent redundant information. To tackle the above problems, this paper treats images as graph structures and introduces a simple contrastive vision GNN (SC-ViG) architecture for remote sensing segmentation. Specifically, we construct a node-masked and edge-masked graph view to obtain an optimal graph structure representation, which can adaptively learn whether to mask nodes and edges. Furthermore, this paper innovatively introduces information bottleneck theory into graph contrastive learning to maximize task-related information while minimizing task-independent redundant information. Finally, we replace the convolutional module in UNet with the SC-ViG module to complete the segmentation and classification tasks of remote sensing images. Extensive experiments on publicly available real datasets demonstrate that our method outperforms state-of-the-art remote sensing image segmentation methods.
\end{abstract}
	
	
\section{Introduction}

\begin{figure}
	\centering
	\includegraphics[width=1\linewidth]{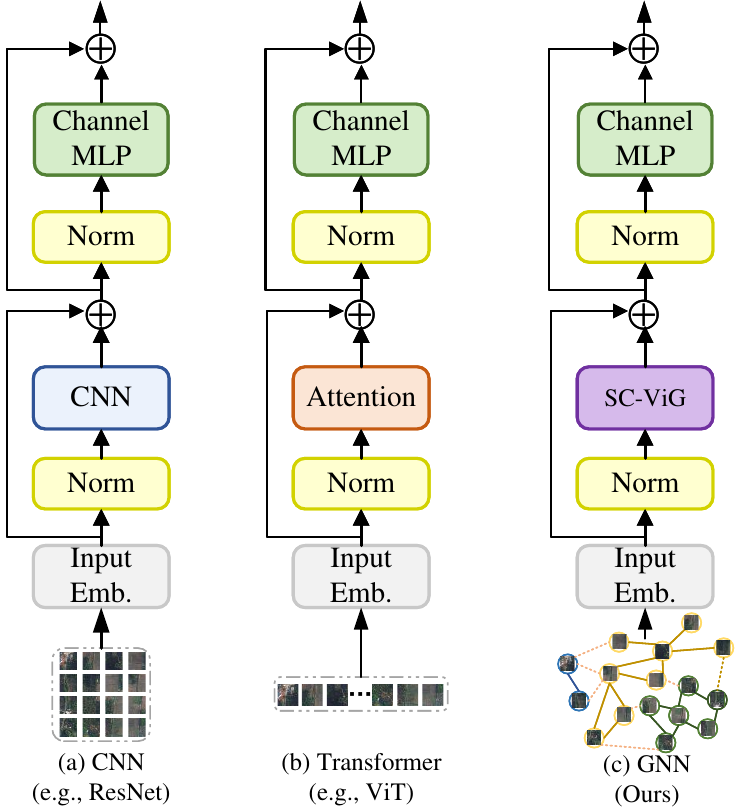}
	\caption{Illustrative examples of different modeling approaches for an image. (a) CNNs view images as regular grid structures (i.e., squares and rectangles). (b) Transformer treats images as a continuous sequence structure. (c) We believe that both sequence structure and grid structure are special cases of graph structure, and graph structure can flexibly model regular and irregular objects. We thus view images as graph structures.}
	\label{fig:1}
\end{figure}

\subsection{Motivation}
\IEEEPARstart{R}{emote} sensing segmentation has been widely developed in a variety of scenarios including land cover mapping, environmental protection, and road information extraction, which require high-quality feature representations to be learned from irregular objects (e.g., roads, trees, etc). In recent years, thanks to the powerful modeling ability for image data, convolutional neural network (CNN) and Transformer with attention module have provided an effective way to extract the underlying visual features and multi-scale features of images and exhibit guaranteed performance on remote sensing segmentation \cite{shou2022object}. 

Although encouraging segmentation performance has been achieved, we argue that CNN-based and Transformer-based remote sensing segmentation models suffer from a limits. Taking Fig. 1 as an example, Fig. 1(a) shows the CNN-based image modeling method, which treats the image as a regular grid structure. Fig. 1(b) shows the Transformer-based image modeling method, which regards the image as a continuous sequence structure. Both of the above methods are unable to model irregular objects. As shown in Figure 1(c), we argue that both grid and sequence structures are special cases of graph structures, and that GNN-based approaches  \cite{10113198}, \cite{10314020}, \cite{yin2023coco}, \cite{10.1145/3503161.3548012},, \cite{shou2022conversational} are capable of modeling data in non-Euclidean spaces. Therefore, we propose a GNN-based remote sensing image modeling method for multi-scale feature extraction of irregular objects. 

Recent advances in graph contrastive representation learning have demonstrated that it can improve model convergence and improve model robustness. Nevertheless, the existing methods suffer from two limitations. First, most existing methods perform feature augmentation by randomly masking graph views to obtain better node representations. However, randomly masking nodes and edges may be too random, which destroys the expressive ability of the semantic information of the original graph. Second, most existing methods generate multiple graph contrastive views and obtain consistent node representations by maximizing the mutual information between them, which may force the model to learn task-independent semantic information. We argue that a good feature augmentation method should minimize task-independent redundant information.

To address the aforementioned issue, we propose a simple contrastive vision GNN (SC-ViG) architecture for remote sensing segmentation, which consists of key steps, i.e., an adaptive feature augmentation module and a graph contrastive learning via information bottleneck module.

First, we introduce a learnable graph contrastive view to adaptively learn whether to mask nodes and edges to improve the node representation ability of the original graph, which is optimized together with downstream remote sensing segmentation and classification in an end-to-end learning manner. The intuition behind adaptive masking is to randomly mask nodes and edges without a policy regardless of the degree of the corresponding nodes. However, GCNs aggregate the information of surrounding neighbor nodes through the message passing mechanism, which makes it easy for GCNs to reconstruct the feature information of popular nodes, but it is difficult to reconstruct the feature information of isolated nodes with low degrees. These adaptive masking generated graph contrastive view  increases the ability against imbalanced learning for remote sensing segmentation.

Second, we propose to integrate different graph-contrastive views into compact representations for downstream remote sensing segmentation tasks, which can further improve the feature representation capabilities of nodes. Recent advances have shown that downstream performance can be improved by fusing complementary semantic information between different views. Therefore, we argue that maximizing the mutual information (MI) between graph contrastive views forces a consistent representation of the graph structure, which leads the model to capture task-independent redundant information. Inspired by the information bottleneck (IB) theory, we use it to minimize the MI between the original graph and the generated contrastive view while preserving task-relevant semantic information. Through the above approach, the model can jointly learn complementary semantic information between different views.

\subsection{Our Contributions}
Compared with previous work, the contributions of this paper are summarized as follows.

\begin{enumerate}
	\item We propose a simple contrastive vision GNN (SC-ViG) architecture for remote sensing segmentation, which enables flexible modeling of irregular objects
	
	\item We introduce a novel graph contrastive learning approach to optimize node representations by adaptively masking nodes and edges, which improves the representation ability of graph structure.
	
	\item We innovatively embed the information bottleneck theory into the graph contrastive learning method, which can effectively eliminate redundant information while preserving task-related information.
	
	\item Extensive experiments demonstrate that our method outperforms the state-of-the-art on three publicly available datasets.
\end{enumerate}
	
\section{Related Work}

\subsection{CNN, and Transformer for Remote Sensing Segentation}
The early mainstream network architecture for remote sensing segmentation extracts visual features of images by using CNN. The earliest remote sensing image segmentation methods based on CNN are all evolved from FCN (\cite{long2015fully}, \cite{wu2022hg}, etc) and UNet (e.g., \cite{zhou2018unet++}, \cite{cao2022swin}, etc). UNet extracts the context and location information of the image by designing a U-shaped structure based on the encoder and decoder, where both of them are composed of convolutional layers, skip connections, and pooling layers. FCN extracts image features through several convolutional layers and then connects a deconvolutional layer to obtain a feature map of the same size as the raw image, so as to predict the image pixel by pixel. However, both FCN and UNet algorithms need to down-sample to continuously expand the receptive field when extracting image features, which leads to the loss of image position information. To alleviate the problem of information loss caused by the downsampling operation, the DeepLab series \cite{liang2015semantic}, uses hole convolution to increase the receptive field to obtain multi-scale feature information. The HRNet proposed by Wang et al. \cite{wang2020deep} achieves high-resolution semantic segmentation by extracting feature maps of different resolutions and recovering high-resolution feature maps.

Transformer \cite{vaswani2017attention} is widely used in the image processing field because of its powerful global information processing capabilities. ViT proposed by Dosovitskiy et al. \cite{dosovitskiy2020image} applied the Transformer architecture to CV for the first time, and she used the attention to extract global visual features. Since the complexity of the attention is $O(n^2)$, this leads to a very large number of parameters in the model, and the model is difficult to train. To solve the above problems, Liu et al. \cite{liu2021swin} proposed Swin-Transformer, which improves the problem of high model complexity through a hierarchical attention mechanism. The Wide-Context Transformer proposed by Ding et al. \cite{9759447} extracts global context information by introducing a Context Transformer while using CNN to extract features. Zhang et al. \cite{zhang2022transformer} extract multi-scale contextual features by combining Swin-Transformer and dilated convolutions, and use a U-shaped decoder to achieve image semantic segmentation.

\begin{figure*}
	\centering
	\includegraphics[width=1\linewidth]{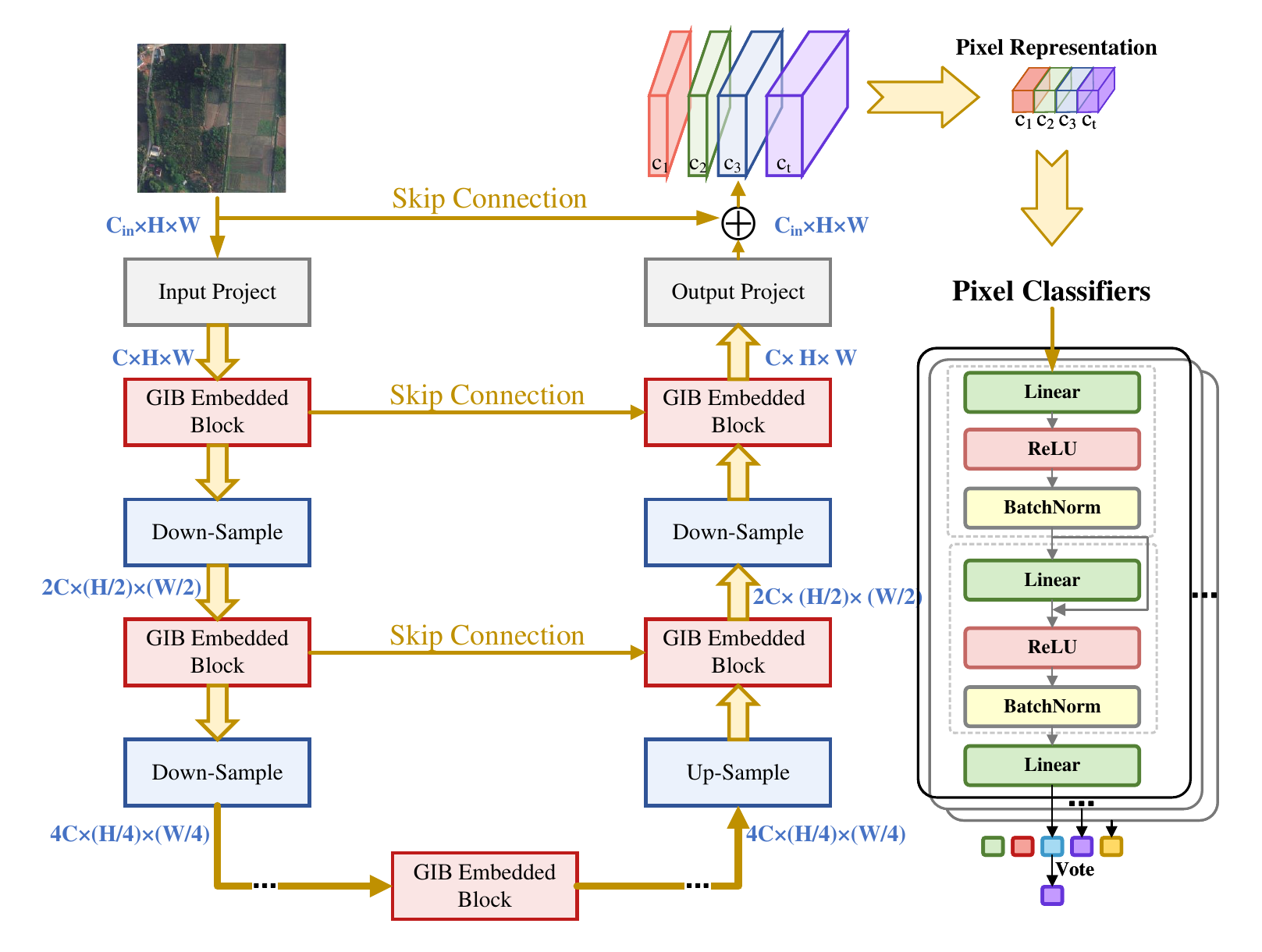}
	\caption{The architecture of the proposed GraphUNet method. Specifically, we first divide the image into patches and construct it as a graph. Then we replace the convolutional block in UNet with our GCN Block and use the constructed graph as the input. Finally, we build a MLP to classify pixels.}
	\label{fig:graphunet}
\end{figure*}

\subsection{Graph Neural Networks}
Kipf et al. \cite{kipf2016semi} were the first to propose graph convolutional neural networks. In recent years, spatial-based GCNs and spectral-based GCNs have started to receive widespread attention, and they are applied to graph-structured data (e.g., social networks \cite{mo2022simple} and citation networks \cite{wen2022trend}, etc.).

In recent years, graph neural network (GNN) has received extensive attention from researchers due to its powerful feature extraction capabilities, and it has been widely used in action recognition \cite{hao2021hypergraph}, point cloud analysis \cite{shi2020point} and other fields \cite{han2022vision}. GNN can flexibly model irregular objects and extract global location feature information. In the remote sensing segmentation field, Saha et al. \cite{9706461} use GNN to aggregate and label unlabeled data to improve the ability of the model to approach the target domain. 

\subsection{Graph Contrastive Learning}
Graph contrastive learning (GCL) aims to learn compact representations of nodes or subgraphs in graph data, emphasizing similarities within the same graph and differences between different graphs. GCL has been applied in many fields, including social network analysis, drug discovery, image analysis, etc. For example, in social networks, similarities between users can be discovered through GCL, and in drug discovery, potential drug similarities can be mined by contrastiving molecular structures.

In recent research, DGI \cite{velivckovic2018deep} and InfoGraph \cite{sun2019infograph} obtain compact representations of graphs or nodes by maximizing the mutual information (MI) between different augmented views. MVGRL \cite{hassani2020contrastive} argues that it can achieve optimal feature representation by contrastiving first-order neighbor nodes and performing node diffusion to maximize the MI between subgraphs. GraphCL \cite{you2020graph} constructs four types of augmented views and maximizes the MI between them. GraphCL enables better generalization performance on downstream tasks. However, GraphCL requires complex manual feature extraction. We argue that a good contrast-augmented view should be structurally heterogeneous while semantically similar, while previous research work maximizes the mutual information between nodes, which may lead to overfitting of the model.

\section{Approach}
In this section, we illustrate the construction of graph-structured data from images, and introduce the GCL architecture with the information bottleneck to learn to extract global information locations of images.

\subsection{Structure Flow}
Our main goal is to design an efficient modeling paradigm for global location information extraction of irregular objects, detailed in Fig. \ref{fig:graphunet}. For a given remote sensing image $(H \times W \times 3)$, we first divide it into $M$ patches. Then we map each image patch to a $D$-dimensional feature space $x_i \in \mathbb{R}^D$, and obtain the feature vector $X$. We consider $X$ to be a node in the graph, i.e., $V = \{v_1, v_2, , v_N\}$. For node $v_i$, we use the KNN algorithm to find its $K$ neighbors $N (v_i)=\{v_i^1, v_i^2, \ldots, v_i^K\}$. For $v_j \in N (v_i)$, we connect an edge  $e_{ji}$ from $v_j$ to $v_i$. Through the above process, we get a directed graph $G = (V, E)$. Follow UNet's network architecture design, feature embedding for images uses $N$ encoders for feature encoding. Each stage consists of a GIB Embedding block (GE), a skip connection module and a downsampling layer. GE Block utilizes the inherent flexible modeling of non-Euclidean distance in the graph structure, follows the global modeling rules of node aggregation, and customizes the global position information interaction of the image. We downsample the feature maps with a $3 \times 3$ kernel. Similarly, the decoder stage consists of the proposed GE block and an upsampling layer to decode and reconstruct features. To ensure the effective utilization of information and the depth of network training, the decoder input of each stage is connected with the output of the encoder of the same stage. Finally, a convolutional layer is applied to generate the segmented image $S \in C_{in} \times H \times W$, which is predicted pixel by pixel.

\subsection{GCN Embedded Block}
The advantages of using a graph structure to model images are as follows: 1) The graph can flexibly handle data with non-Euclidean distances. 2) Compared with regular grid or sequence structures, graphs can model irregular objects while eliminating redundant information, and remote sensing images are mostly irregular objects. 3) The graph structure establishes the connection between objects (e.g., roads, trees, etc) through the connection between nodes and edges.

Specifically, for an input image feature $X$, we first construct a directed graph $G = G(x)$. To obtain the global location information of the image and update node features, we use graph convolution operations to aggregate and update node features. The formula is defined as follows:

\begin{equation}
	\begin{aligned}
		{G}^{\prime} & =F({G}, \mathcal{W}) \\
		& =\operatorname{Update}\left(\text { Aggregate }\left(G, W_{\text {agg }}\right), W_{\text {update }}\right) \\
		& =\text { LeakyReLU }\left(\sum _ { r \in \mathcal { R } } \sum _ { j \in \mathcal { N } _ { i } ^ { r } } \frac { 1 } { | \mathcal { N } _ { i } ^ { r } | } \left(\omega_{i j}^{(l)} W_{\theta_1}^{(l)} x_j^{(l)}\right.\right. \\
		& \left.\left.+\omega_{i i}^{(l)} W_{\theta_2}^{(l)} x_i^{(l)}\right)\right)
	\end{aligned}
\end{equation}
where $W_{agg}$, $W_{update}, W_{\theta_1}^{(l)}, W_{\theta_2}^{(l)}$ is learnable weights, $w_{ij}$ is the edge weight between node $i$ and node $j$, and its formula is defined as follows:
\begin{equation}
	\begin{aligned}
		\omega_{i j}^{(l+1)} & =\operatorname{softmax}\left(W^{(l)}[x_i^{(l)} \oplus x_j^{(l)}]\right) \\
		& =\frac{\exp \left[x_i^{(l)} \oplus x_j^{(l)}\right]}{\sum_{\eta \in \mathcal{N}_i} \exp \left[x_i^{(l)} \oplus x_j^{(l)}\right]},
	\end{aligned}
\end{equation}

\begin{figure*}
	\centering
	\includegraphics[width=1\linewidth]{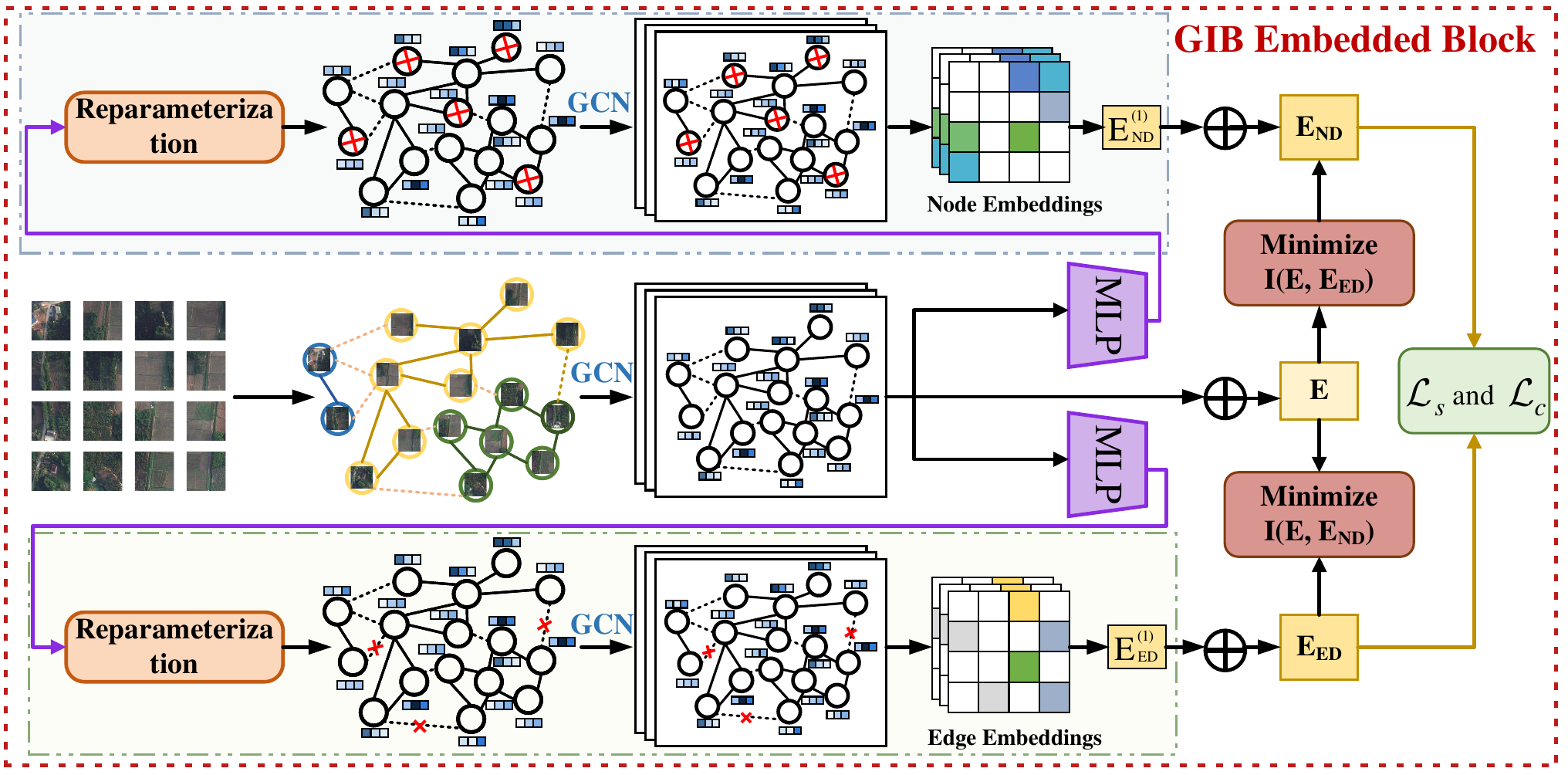}
	\caption{The overview the GCN Embedded Block framework. We synthesize node-mask and edge-mask views to obtain better node representations. Specifically, we introduce information bottleneck theory in multiple graph comparison views to maximize feature information related to node classification tasks while minimizing redundant information of nodes.}
	\label{fig:3}
\end{figure*}

To capture the location information of key regions in the image, we further introduce a multi-head attention mechanism to update node features. The format is defined as follows:

\begin{equation}
	\mathbf{x}_i^{\prime}=[\text { head}^1 W_{\text {update }}^1, \text { head}^2 W_{\text {update}}^2, \cdots, \text { head}^h W_{\text {update}}^h]
\end{equation}
where $h$ represents the number of multi heads, we set $h=4$.

We introduce the residual idea, and project node features to the same domain through a linear layer, which can help restore structural features and global position information. In addition, we also insert the LeakyReLU non-activation function to improve the nonlinear fitting ability of the model. The formula is expressed as follows:
\begin{equation}
	Y=\text {LeakyReLU}\left(\operatorname{GraphConv}\left(X W_{\text {in }}\right)\right) W_{\text {out }}+X
\end{equation}

To improve the feature transformation ability of nodes and alleviate the over-smoothing phenomenon of GCN, we use feed-forward network (FFN) to perform feature mapping on each node again. The formula of FFN is defined as follows:
\begin{equation}
	Y'=\text{LeakyReLU}\left(Y W_1\right) W_2+Y
\end{equation}
where $W_1$ and $W_2$ are the learnable parameters.

\subsection{Graph Information Bottleneck}
The principle of graph information bottleneck (GIB) is to introduce information bottleneck (IB) on the basis of GCL to perform contrastive learning between nodes or graphs. It forces the node representation $Z_X$ to minimize the task-irrelevant information $\mathcal{D}$ and maximize the information $Y$ relevant to the downstream tasks.

Specifically, we follow the local dependency assumption for graph-structured data: for a given node $v$, node $v$’s first-order neighbor node data are related to node $v$, while the rest of the graph’s data are independent and identically distributed with node $v$. The hypothesis space represented by nodes can be constrained according to local dependency assumptions, which reduces the difficulty of GIB optimization. We assume that $\mathbb{P}(Z_X|\mathcal{D})$ represents modeling the correlation between node features hierarchically. In each iteration $l$, the representation of each node is optimized by aggregating surrounding neighbor node information and graph structure information $Z^{(l)}_A$. Therefore, the optimization goal of GIB is defined as follows:
\begin{equation}
	\begin{aligned}
	&\min_{\mathbb{P}(Z_X^{(L)}|\mathcal{D})\in\Omega}\mathrm{GIB}_\beta(\mathcal{D},Y;Z_X^{(L)}) \\ &\triangleq \left[-I(Y;Z_X^{(L)})+\beta I(\mathcal{D};Z_X^{(L)})\right]
	\end{aligned}
	\label{tab:eq6}
\end{equation}
where $\Omega$ conforms to the representation space of Markov chain probability dependence within a given data set $\mathcal{D}$, $I(,)$ represents mutual information between feature vectors, $Z_X^{(L)}$ represents the feature representations of the nodes, and $\beta$ is the balance coefficient. In Eq. \ref{tab:eq6}, the model only needs to optimize two distributions, i.e., $\mathbb{P}(Z_A^{(l)}|Z_X^{(l-1)},A)$, and $\mathbb{P}(Z_X^{(l)}|Z_X^{(l-1)},Z_A^{(l)})$, where $Z_A^{(l)}$ is the graph structure information.

However, in Eq. 6, calculating the mutual information $I(Y;Z_X^{(L)})$ and $I(\mathcal{D};Z_X^{(L)})$ is a difficult estimation problem. Therefore, we follow the IB criterion to introduce variational bounds on $I(Y;Z_X^{(L)})$ and $I(\mathcal{D};Z_X^{(L)})$ to effectively perform parameter optimization. We give the upper and lower bounds of $I(\mathcal{D};Z_X^{(L)})$ and $I(Y;Z_X^{(L)})$ as shown in Theorems 1 and 2 respectively.

\textbf{Theorem 1.} For any class distribution given $\mathbb{Q}_1(Y_v|Z_{X,v}^{(L)})$ for $v \in V$ and $\mathbb{Q}_2(Y)$ in a graph, we can obtain a theoretical lower bound for $I(Y;Z_X^{(L)})$:
\begin{equation}
	\begin{aligned}
	I(Y;Z_X^{(L)}) & \geq1+\mathbb{E}\left[\log\frac{\prod_{v\in V}\mathbb{Q}_1(Y_v|Z_{X,v}^{(L)})}{\mathbb{Q}_2(Y)}\right] \\ &+\mathbb{E}_{\mathbb{P}(Y)\mathbb{P}(Z_X^{(L)})}\left[\frac{\prod_{v\in V}\mathbb{Q}_1(Y_v|Z_{X,v}^{(L)})}{\mathbb{Q}_2(Y)}\right]
	\end{aligned}
\end{equation}

\textbf{Theorem 2.} For any given node feature distribution $\mathbb{Q}(Z_X^{(l)})$ and graph structure information distribution $\mathbb{Q}(Z_A^{(l)})$, we use Markov chain dependence to get the upper bound of $I(Y;Z_X^{(L)})$ as follows:
\begin{equation}
	\begin{aligned}
	I(\mathcal{D};Z_X^{(L)}) & \leq I(\mathcal{D};\{Z_X^{(l)}\}_{l\in S_X}\cup\{Z_A^{(l)}\}_{l\in S_A}) \\
	& \leq\sum_{l\in S_A}\mathrm{AIB}^{(l)}+\sum_{l\in S_X}\mathrm{XIB}^{(l)}
	\end{aligned}
\end{equation}
where $l \in \{S_X, S_A\}$, and 

\begin{equation}
	\begin{aligned}
	\mathrm{AIB}^{(l)}=\mathbb{E}\left[\log\frac{\mathbb{P}(Z_A^{(l)}|A,Z_X^{(l-1)})}{\mathbb{Q}(Z_A^{(l)})}\right],\\ \mathrm{XIB}^{(l)}=\mathbb{E}\left[\log\frac{\mathbb{P}(Z_X^{(l)}|Z_X^{(l-1)},
	Z_A^{(l)})}{\mathbb{Q}(Z_X^{(l)})}\right]
	\end{aligned}
\end{equation}
where $\mathrm{AIB}$ and $\mathrm{XIB}$ represents the adjacency matrix features and  the node features obtained using the IB criterion, respectively.

To use GIB, we optimize $\mathbb{P}(Z_A^{(l)}|Z_X^{(l-1)},A)$ and $\mathbb{P}(Z_X^{(l)}|Z_X^{(l-1)},Z_A^{(l)})$  given a theoretical upper and lower bound. Next, we will specify the optimization goals of GIB.

\textbf{Objective for training.} To update model parameters in GIB, we need to calculate the theoretical boundary of GIB in \ref{tab:eq6}. Specifically, we use a uniform distribution to optimize the classification problem: $Z_A \sim \mathbb{Q}(Z_A),Z_{A,v}=\cup_{t=1}^{\mathcal{T}}\{u\in V_{vt}|u\overset{\mathrm{iid}}{\operatorname*{\sim}}\operatorname{Cat}(\frac1{|V_{\upsilon t}|})\}$. Therefore, we can obtain an estimate of $\mathrm{AIB}^{(l)}$ as follows:
\begin{equation}
	\widehat{\mathrm{AIB}}^{(l)}=\mathbb{E}_{\mathbb{P}(Z_A^{(l)}|A,Z_X^{(l-1)})}\left[\log\frac{\mathbb{P}(Z_A^{(l)}|A,Z_X^{(l-1)})}{\mathbb{Q}(Z_A^{(l)})}\right]
\end{equation}

$\mathrm{AIB}^{(l)}$ can be formally defined as follows:
\begin{equation}
	\begin{aligned}\widehat{\operatorname{AIB_C}}^{(l)}=\sum_{v\in V,t\in[\mathcal{T}]}\operatorname{KL}(\operatorname{Cat}(\phi_{vt}^{(l)})||\operatorname{Cat}(\frac1{|V_{vt}|}))
	\end{aligned}
\label{eq:11}
\end{equation}

For the estimation of XIB, we use a learnable Gaussian distribution to set $\mathbb{Q}(Z_X^{(l)})$. Specifically, for a given node $v$, ${Z_X}\sim\mathbb{Q}(Z_X)$, we assume $Z_{X,v}\sim\sum_{i=1}^mw_i\text{Gaussian}(\mu_{0,i},\sigma_{0,i}^2)$. Therefore $\widehat{\mathrm{XIB}}^{(l)}$ is formally defined as follows:
\begin{equation}
	\begin{aligned}
	\widehat{\mathrm{XIB}}^{(l)} & =\log\frac{\mathbb{P}(Z_X^{(l)}|Z_X^{(l-1)},Z_A^{(l)})}{\mathbb{Q}(Z_X^{(l)})} \\ & =\sum_{v\in V}\left[\log\Phi(Z_{X,v}^{(l)};\mu_v,\sigma_v^2)\right.
	\\ & \left.-\log(\sum_{i=1}^{m}w_i\Phi(Z_{X,v}^{(l)};\mu_{0,i},\sigma_{0,i}^2))\right]
	\end{aligned}
\label{eq:12}
\end{equation}
where $\mu_{0,i},\sigma_{0,i}, w_i$ are the learnable.

Combining Eqs. \ref{eq:11} and \ref{eq:12}, we can estimate $I(\mathcal{D};Z_X^{(L)})$ as follows:
\begin{equation}
	I(\mathcal{D};Z_X^{(L)})\to\sum_{l\in S_A}\widehat{\mathrm{AIB}}^{(l)}+\sum_{l\in S_X}\widehat{\mathrm{XIB}}^{(l)}
	\label{eq:13}
\end{equation}

We use cross entropy to estimate $I(Y;Z_X^{(L)})$ as follows:
\begin{equation}
	I(Y;Z_X^{(L)})\to-\sum_{v\in V}\text{Cross-Entropy}(Z_{X,v}^{(L)}W_{\mathrm{out}};Y_v)
	\label{eq:14}
\end{equation}

By combining Eqs. \ref{eq:13} and \ref{eq:14}, we can get the optimization objective of GIB.

\subsection{Instantiating GIB-RSS}
After detailing the optimization principles of GIB, we will explain the GIB-RSS architecture we designed as shown in Fig. \ref{fig:3}.

\textbf{Node-Masking View.} To improve the feature representation ability of nodes in the learning process, we perform learnable node masking before each information aggregation and feature update of GCN. The formula for the node mask view we created is as follows:
\begin{equation}
	\mathcal{G}_{N D}^{(l)}=\left\{\left\{v_i \odot \eta_i^{(l)} \mid v_i \in \mathcal{V}\right\}, \mathcal{E}, \mathcal{R}, \mathcal{W}\right\},
\end{equation}
where $\eta_i^{(l)} \in \{0,1\}$ is sampled from a parameterized Bernoulli distribution $Bern(\omega_i^l)$, and $\eta_i^{(l)} = 0$ represents masking node $v_i$, $\eta_i^{(l)} = 1$ represents keeping node $v_i$.

\textbf{Edge-masking View.} The goal of the edge-masking view is to generate an optimized graph structure, and the formula is defined as follows:
\begin{equation}
	\mathcal{G}_{E D}^{(l)}=\left\{\mathcal{V},\left\{e_{i j} \odot \eta_{i j}^{(l)} \mid e_{i j} \in \mathcal{E}, \mathcal{R}, \mathcal{W}\right\}\right\},
\end{equation}
where $\eta_
{ij}^{(l)} \in \{0,1\}$ is also sampled from a parameterized Bernoulli distribution $Bern(\omega_{ij}^l)$, and $\eta_{ij}^{(l)} = 0$ represents perturbating edges $e_{ij}$, $\eta_i^{(l)} = 1$ represents keeping edge $e_{ij}$.

After obtaining the masked node and edge-masking views, we input them into GCN for feature representation to obtain optimized multi-views. The formula is defined as follows:
\begin{equation}
	\begin{aligned}
		&\mathbf{E}_{N D}^{(l)}=GraphConv\left(\mathbf{E}_{N D}^{(l-1)}, \mathcal{G}_{N D}^{(l)}\right), \\ 
		&\mathbf{E}_{E D}^{(l)}=GraphConv\left(\mathbf{E}_{E D}^{(l-1)}, \mathcal{G}_{E D}^{(l)}\right).
	\end{aligned}
\end{equation}
where $GraphConv$ represents the graph convolution operation, and we choose GAT as our graph encoder. $\mathbf{E}_{N D}$ and $\mathbf{E}_{E D}$ represent the node feature representations of node-masking view and edge-masking view respectively, $\mathbf{G}_{N D}$ and $\mathbf{G}_{E D}$ represent node-masking view and edge-masking view respectively.

After obtaining the node mask and edge mask views, we combine Eqs. \ref{eq:13} and \ref{eq:14} to jointly optimize the self-supervised losses $\mathcal{L}_s$ and $\mathcal{L}_c$ as follows:
\begin{equation}
	\begin{aligned}
	\min (\mathcal{L}_s + \mathcal{L}_c) &= I(\mathcal{D}^{(ED)};Z_X^{(ED)}) + I(Y;Z_X^{(ED)}) 
	\\ &+ I(\mathcal{D}^{(ND)};Z_X^{(ND)}) + I(Y;Z_X^{(ND)})
	\end{aligned}
\end{equation}
where $\mathcal{D}^{(ED)}$ and $\mathcal{D}^{(ND)}$ represent the graph structure of the node-masking and edge-masking views respectively, and $Z_X^{(ED)}$ and $Z_X^{(ND)}$ represent the node features of the node-masking and edge-masking views respectively.

\subsection{Model Training}
The cross-entropy loss function is used to measure the difference between the probability distribution of the model output and the distribution of the actual labels, thereby guiding the model to optimize parameters.

\begin{equation}
	\mathcal{L}=-\frac{1}{\sum_{s=1}^{N}c(s)}\sum_{i=1}^{N}\sum_{j=1}^{c(i)}\log\mathcal{P}_{i,j}[y_{i,j}]+\lambda\left\|\theta\right\|_{2}
\end{equation}
where $N$ represents the number of remote sensing images used for training, $c(i)$ represents the number of objects to be segmented in sample $i$, $\mathcal{P}[i,j]$ is the label probability distribution of object $j$ in sample $i$, $y_{i,j}$ is the prediction label of object $j$ in sample $i$, $\lambda$ is the weight attenuation coefficient, and $\theta$ is the trainable network parameter.

\section{Experiments}
In this section, we verify the effectiveness of the proposed GraphUNet on remote sensing image segmentation tasks.

\subsection{Benchmark Datasets Used}
For the GraphUNet model, we use the widely used datasets UAVid \cite{lyu2020uavid}, Vaihingen \cite{li2020scattnet} and Potsdam \cite{boguszewski2021landcover} datasets for experimental evaluation. The UAVid dataset comes with two spatial resolutions. Specifically, the UAVid dataset contains a total of 420 images, and each image is cropped to a size of 1024 $\times$ 1024. The Vaihingen dataset consists of 33 images with a spatial resolution of 2494 $\times$ 2064. Each image is cropped to 1024 $\times$ 1024. The Potsdam dataset contains 38 image patches with a spatial resolution of 6000 $\times$ 6000, and we crop the original image size to 1024 $\times$ 1024. The LoveDA dataset contains 5, 987 high-resolution remote sensing images with size 1024×1024, 2, 522 images are used for training, 1, 669 images are used for validation, and 1, 796 images are used for testing. The data information of the dataset is shown in Table \ref{table:info}.

\begin{table}[htbp]
	\centering
	\caption{The division of the train set, val set and test set in the benchmark dataset and the resolution information of the image.}
	\label{table:info}
	\renewcommand\arraystretch{1.5}
	\setlength{\tabcolsep}{7pt}{
		\begin{tabular}{ccccc}
			\toprule
			Datasets                    & Resolutions                           & Train & Test & Val  \\ \midrule
			UAVid                       & 3840 $\times$ 2160/4096 $\times$ 2160 & 200   & 150  & 70   \\
			Vaihingen                 & 2494 $\times$ 2064                    & 15    & 17   & 1   \\
			Potsdam & 6000 $\times$ 6000                    & 22  & 14 & 1  \\
			LoveDA   &  1024 $\times$ 1024      &   2,522    &   1,796     & 1,669
			\\ \bottomrule
	\end{tabular}}
\end{table}

\begin{table*}[!t]
	\centering
	\caption{Experimental results of our method and SOTA methods on the UAVid dataset. The optimal values in columns are shown in bold.}
	\label{tab:exp1}
	\renewcommand\arraystretch{1.5}
	\setlength{\tabcolsep}{8.6pt}{
		\begin{tabular}{lcccccccccc}
			\toprule
			\multicolumn{1}{l}{Methods} & Backbone  & Clutter & Building & Road & Tree & Vegetation & MovingCar & StaticCar & Human & mIoU \\ \midrule
			MSD                         & -         & 56.8    & 79.6     & 73.9 & 73.9 & 56.1       & 63.2      & 31.8      & 20.0  & 56.9 \\
			CANet                       & -         & 65.8    & 87.0     & 61.9 & 78.8 & 77.9       & 48.0      & \textbf{68.5}      & 20.0  & 63.5 \\
			DANet                       & ResNet  & 65.1    & 86.2     & 78.0 & 77.9 & 60.9       & 60.0      & 47.1      & 8.9   & 60.5 \\
			SwiftNet                    & ResNet  & 63.9    & 84.9     & 61.3 & 78.3 & 76.4       & 51.2      & 62.4      & 15.8  & 61.8 \\
			BiSeNet                     & ResNet  & 64.5    & 85.8     & 61.0 & 78.1 & 77.1       & 48.8      & 63.2      & 17.4  & 62.0 \\
			MANet                       & ResNet  & 64.4    & 85.1     & 77.9 & 77.4 & 60.5       & 67.5      & 53.4      & 14.6  & 62.6 \\
			ABCNet                      & ResNet  & 67.3    & 86.1     & 81.5 & 79.7 & 63.3       & 69.2      & 48.3      & 13.6  & 63.6 \\
			Segmenter                   & ViT-Tiny  & 63.7    & 85.2     & 80.1 & 77.0 & 58.1       & 58.4      & 35.3      & 13.9  & 59.0 \\
			SegFormer                   & MiT-B1    & 67.3    & 87.3     & 79.8 & 80.1 & 62.7       & 71.7     & 52.7      & 29.3  & 66.3 \\
			BANet                       & ResT-Lite & 65.9    & 86.0     & 81.2 & 79.1 & 61.9       & 68.7      & 52.4      & 20.5  & 64.4 \\
			BoTNet                      & ResNet  & 65.4    & 85.1     & 79.1 & 78.4 & 61.2       & 66.3      & 52.0      & 23.1  & 63.8 \\
			CoaT                        & CoaT-Mini & 68.9    & \textbf{89.1}     & 79.8 & 80.4 & 61.7       & 69.5      & 60.2      & 19.1  & 66.1 \\
			UNetFormer                  & ResNet  & 67.7    & 86.4     & 82.0 & 81.2 & 64.1       & \textbf{74.0}      & 55.8      & \textbf{30.9}  & 67.8 \\ 
			GIB-RSS &-      & \textbf{71.2}  & 89.0    & \textbf{83.0}      & \textbf{81.9}    & \textbf{79.7}    & 70.6  & 
			59.3 & 29.9        & \textbf{70.6} \\
			\bottomrule
	\end{tabular}}
\end{table*}

\begin{table*}[htbp]
	\centering
	\caption{Experimental results of our method and SOTA lightweight methods on the Vaihingen dataset. The optimal values in columns are shown in bold.}
	\label{tab:exp2}
	\renewcommand\arraystretch{1.5}
	\setlength{\tabcolsep}{12.7pt}{
		\begin{tabular}{lccccccccc}
			\toprule
			Methods    & Backbone  & Imp.suf. & Building & Lowveg. & Tree & Car  & MeanF1 & OA   & mIoU \\ \midrule
			DABNet     & -         & 88.0     & 89.1     & 73.9    & 85.0 & 59.9 & 79.2   & 83.9 & 69.9 \\
			ERFNet     & -         & 88.7     & 89.8     & 76.2    & 86.1 & 54.0 & 79.0   & 86.2 & 70.3 \\
			BiSeNet    & ResNet  & 88.7     & 90.7     & 81.0    & 87.1 & 72.9 & 84.1   & 86.6 & 76.3 \\
			PSPNet     & ResNet  & 88.8     & 92.9     & 81.8    & 88.1 & 44.2 & 79.2   & 88.0 & 76.1 \\
			DANet      & ResNet  & 89.7     & 94.1     & 81.9    & 86.9 & 44.6 & 79.4   & 87.6 & 69.6 \\
			FANet      & ResNet  & 91.2     & 94.1     & 83.1    & 88.7 & 72.0 & 85.8   & 90.0 & 76.0 \\
			EaNet      & ResNet  & 92.1     & 94.7     & 82.9    & 88.8 & 80.4 & 87.8   & 90.0 & 78.5 \\
			ShelfNet   & ResNet  & 92.1     & 94.8     & 84.1    & 88.9 & 78.0 & 87.6   & 90.1 & 77.9 \\
			MARsU-Net  & ResNet  & 91.8     & 94.8     & 84.1    & 89.0 & 78.2 & 87.6   & 89.7 & 78.9 \\
			SwiftNet   & ResNet  & 91.9     & 95.1     & 83.6   & 89.6 & 80.6 & 88.2   & 90.0 & 79.2 \\
			ABCNet     & ResNet  & 93.1     & 94.8     & 84.8    & 90.0 & 84.7 & 89.5   & 91.2 & 81.0 \\
			BoTNet     & ResNet  & 90.0     & 91.6     & 82.4    & 89.1 & 72.4 & 85.1   & 87.8 & 74.2 \\
			BANet      & ResT-Lite & 91.6     & 94.8     & 84.0    & 90.3 & 87.2 & 89.6   & 90.5 & 81.4 \\
			Segmenter  & ViT-Tiny  & 90.1     & 92.6     & 80.7    & 90.3 & 68.2 & 84.4   & 87.6 & 73.7 \\
			UNetFormer & ResNet  & 93.1     & 94.9     & 85.2    & \textbf{90.8} & 88.2 & 90.4   & 90.7 & 83.2 \\ 
			GIB-RSS & -  & \textbf{94.7}  & \textbf{96.8}     & \textbf{86.8}     & 89.9    & \textbf{91.5} & \textbf{91.8} & 
			\textbf{92.9} & \textbf{85.3} \\
			\bottomrule
	\end{tabular}}
\end{table*}

\begin{table*}[htbp]
	\centering
	\caption{Experimental results of our method and SOTA lightweight methods on the Potsdam dataset. The optimal values in columns are shown in bold.}
	\label{tab:exp3}
	\renewcommand\arraystretch{1.5}
	\setlength{\tabcolsep}{13pt}{
		\begin{tabular}{lccccccccc}
			\toprule
			Methods     & Backbone  & Imp.suf. & Building & Lowveg. & Tree & Car  & MeanF1 & OA   & mIoU \\ \midrule
			ERFNet      & -         & 89.9     & 92.6     & 81.0    & 76.4 & 90.8 & 86.2      & 84.6 & 75.9 \\
			DABNet      & -         & 90.0     & 92.7     & 83.3    & 81.9 & 93.1 & 88.0   & 87.2 & 79.4 \\
			BiSeNet     & ResNet  & 90.2     & 94.6     & 85.5    & 86.2 & 92.7 & 89.8   & 88.2 & 81.7 \\
			EaNet       & ResNet  & 92.0     & 95.7     & 84.3    & 85.7 & 95.1 & 90.6   & 88.7 & 83.4 \\
			MARsU-Net   & ResNet  & 91.4     & 95.6     & 85.8    & 86.6 & 93.3 & 90.5   & 89.0 & 83.9 \\
			DANet       & ResNet  & 91.0     & 95.6     & 86.1    & 87.6 & 84.3 & 88.9   & 89.1 & 80.3 \\
			SwiftNet    & ResNet  & 91.8     & 95.9     & 85.7    & 86.8 & 94.5 & 91.0   & 89.3 & 83.8 \\
			FANet       & ResNet  & 92.0     & 96.1     & 86.0    & 87.8 & 94.5 & 91.3   & 89.9 & 84.2 \\
			ShelfNet    & ResNet  & 92.5     & 95.8     & 86.6    & 87.1 & 94.6 & 91.3   & 89.9 & 84.4 \\
			ABCNet      & ResNet  & 93.5     & 96.9     & 87.9    & 89.1 & 95.8 & 92.7   & 91.3 & 86.5 \\
			Segmenter   & ViT-Tiny  & 90.9     & 94.6     & 84.9    & 84.7 & 89.1 & 88.7   & 89.3 & 81.1 \\
			BANet       & ResT-Lite & 92.6     & 95.8     & 86.5    & 88.9 & 96.2 & 91.9   & 91.7 & 85.7 \\
			SwinUperNet & Swin-Tiny & 92.7     & 96.5     & 88.0    & 88.4 & 95.8 & 91.7   & 91.2 & 86.0 \\
			UNetFormer  & ResNet  & 93.8     & 96.9     & 88.1    & 89.3 & 96.8 & 93.1   & 91.0 & 87.4 \\ 
			GIB-RSS   & -         & \textbf{94.9}     & \textbf{97.9}     & \textbf{88.7}    & \textbf{90.7} & \textbf{97.2} & \textbf{93.9}      & \textbf{93.5}      & \textbf{87.8}    \\
			\bottomrule
	\end{tabular}}
\end{table*}

\subsection{Experimental Settings}
GraphUNet is implemented on NVIDIA A100 GPU with 80G memory using PyTorch framework. For the hyperparameters in the experiments, the paper utilize the AdamW optimizer for gradient updates. The GraphUNet's learning rate (LR) is set to 5e-4 and a cosine learning rate decay is utilized to dynamically adjust the LR. During model training, we use a random flip strategy for data augmentation. For the UAVid dataset, we crop the image size to 1024 $\times$ 1024. For Vaihinge, Potsdam datasets, we crop images to 512 $\times$ 512. When the GraphUNet is trained, we set epoch to 80, and batch size to 32.

\subsection{Evaluation Metrics}
We used multiple evaluation metrics to evaluate the experimental performance of all models, including Overall Accuracy (OA), meanF1, and mIoU. OA, F1 and mIOU reflect the accuracy of remote sensing image segmentation from different angles.

\subsection{Baseline models}
\textbf{MSD:} The Multi-Scale-Dilation (MSD) method proposed by Lyu et al. \cite{lyu2020uavid} achieves image segmentation by using a large-scale pre-trained model to extract multi-scale features of the image.

\textbf{CANet:} The Context Aggregation Network (CANet) proposed by Yang et al. \cite{yang2021real} effectively extracts the spatial information and global information of the image by building a dual-branch CNN and uses an aggregation mechanism to fuse the spatial and global context information.

\textbf{DANet:} The dual attention network (DANet) proposed by Fu et al. \cite{fu2019dual} can achieve the extraction and fusion of global and local semantic information in space and channels.

\textbf{SwiftNet:} SwiftNet proposed by Orsic et al. \cite{orvsic2021efficient} uses a pyramid structure to perform feature fusion of local information. SwiftNet adds regularization terms to constrain the model during the optimization process.

\textbf{BiSeNet:} The Bilateral Segmentation Network (BiSeNet) proposed by Yu et al. \cite{yu2018bisenet} extracts spatial information and high-resolution features by setting small-stride spatial convolution kernels. At the same time, a down-sampling strategy is used to extract contextual information, and a fusion module is designed to achieve effective fusion of information.

\textbf{MANet:} The multi-attention network (MANet) proposed by Li et al. \cite{li2021multiattention} reduces the computational load of the model by building a linear attention module to ensure modeling context dependencies.

\textbf{ABCNet:} The Attention Bilateral Context Network (ABCNet) proposed by Li et al. \cite{li2021abcnet} can lightweightly extract spatial information and contextual information of images.

\textbf{Segmenter:} Segmenter proposed by Strudel et al. \cite{strudel2021segmenter} introduces ViT to realize the modeling of global context information. Unlike CNN, Segmenter can obtain class labels pixel by pixel.

\begin{table*}[!t]
	\caption{Experimental results of our method and state-of-the-art methods on the LoveDA dataset. The optimal values in columns are shown in bold.}
	\label{tab:4}
	\renewcommand\arraystretch{1.5}
	\setlength{\tabcolsep}{7pt}{
		\begin{tabular}{lccccccccccc}
			\hline
			Methods     & Backbone  & Background & Building & Road & Water & Barren & Forest & Agriculture & mIoU & Complexity & Speed \\ \hline
			PSPNet      & ResNet50  & 44.4       & 52.1     & 53.5 & 76.5  & 9.7    & 44.1   & 57.9        & 48.3 & 105.7      & 52.2  \\
			DeepLabV3++ & ResNet50  & 43.0       & 50.9     & 52.0 & 74.4  & 10.4   & 44.2   & 58.5        & 47.6 & 95.8       & 53.7  \\
			SemanticFPN & ResNet50  & 42.9       & 51.5     & 53.4 & 74.7  & 11.2   & 44.6   & 58.7        & 48.2 & 103.3      & 52.7  \\
			FarSeg      & ResNet50  & 43.1       & 51.5     & 53.9 & 76.6  & 9.8    & 43.3   & 58.9        & 48.2 & -          & 47.8  \\
			FactSeg     & ResNet50  & 42.6       & 53.6     & 52.8 & 76.9  & 16.2   & 42.9   & 57.5        & 48.9 & -          & 46.7  \\
			BAnet       & ResNet50  & 43.7       & 51.5     & 51.1 & 76.9  & 16.6   & 44.9   & 62.5        & 49.6 & 52.6       & 11.5  \\
			TransUNet   & ViT-R50   & 43.0       & 56.1     & 53.7 & 78.0  & 9.3    & 44.9   & 56.9        & 48.9 & 803.4      & 13.4  \\
			Segmenter   & ViT-Tiny  & 38.0       & 50.7     & 48.7 & 77.4  & 13.3   & 43.5   & 58.2        & 47.1 & 26.8       & 14.7  \\
			SwinUperNet & Swin-Tiny & 43.3       & 54.3     & 54.3 & 78.7  & 14.9   & 45.3   & 59.6        & 50.0 & 349.1      & 19.5  \\
			DC-Swin     & Swin-Tiny & 41.3       & 54.5     & 56.2 & 78.1  & 14.5   & 47.2   & 62.4        & 50.6 & 183.8      & 23.6  \\
			UNetFormer  & ResNet18  & 44.7       & 58.8     & 54.9 & 79.6  & 20.1   & 46.0   & 62.5        & 52.4 & 46.9       & 115.3 \\ 
			GIB-RSS  & -  & \textbf{45.8}       & \textbf{59.6}     & \textbf{56.4} & \textbf{80.4}  & \textbf{21.2}   & \textbf{48.2}   & \textbf{63.7}        & \textbf{54.1} & \textbf{34.2}       & \textbf{122.1} \\\hline
	\end{tabular}}
\end{table*}

\textbf{SegFormer:} SegFormer proposed by Xie et al. \cite{xie2021segformer} combines Transformer and MLP to extract multi-scale features of images in a hierarchical manner.

\textbf{BANet:} Wang et al. \cite{wang2021transformer} proposed a bilateral perception network (BANet) to extract texture information and boundary information in images in a fine-grained manner. BANet is based on the Transformer pre-training model to achieve information fusion.

\textbf{BoTNet:} The BoTNet proposed by Srinivas et al. \cite{srinivas2021bottleneck} integrates the self-attention mechanism into the ResNet module to extract the global context information of the image.

\textbf{TransUNet:} TransUNet proposed by Chen et al. \cite{chen2021transunet} embeds Transformer's self-attention mechanism into the structure of UNet so that the model can better capture the global relationship of the input image.

\textbf{ShelfNet:} ShelfNet proposed by Zhuang et al. \cite{zhuang2019shelfnet} adopts a multi-resolution processing strategy, which processes input images at different levels. Such a design allows the network to better capture local details in the image while retaining the global information of the image.

\textbf{CoaT:} CoaT proposed by Xu et al. \cite{xu2021co} adopts a co-scaling mechanism to maintain the integrity of the Transformers encoder branch at different scales and provide rich multi-scale and contextual information.

\textbf{UNetFormer:} UNetFormer proposed by Wang et al. \cite{wang2022unetformer} introduces the Transformer mechanism based on UNet. In UNetFormer, Transformer is used to better capture the global contextual information in the image and improve the model's ability to understand the overall structure.

\begin{figure*}[htbp]
	\centering
	\includegraphics[width=1\linewidth]{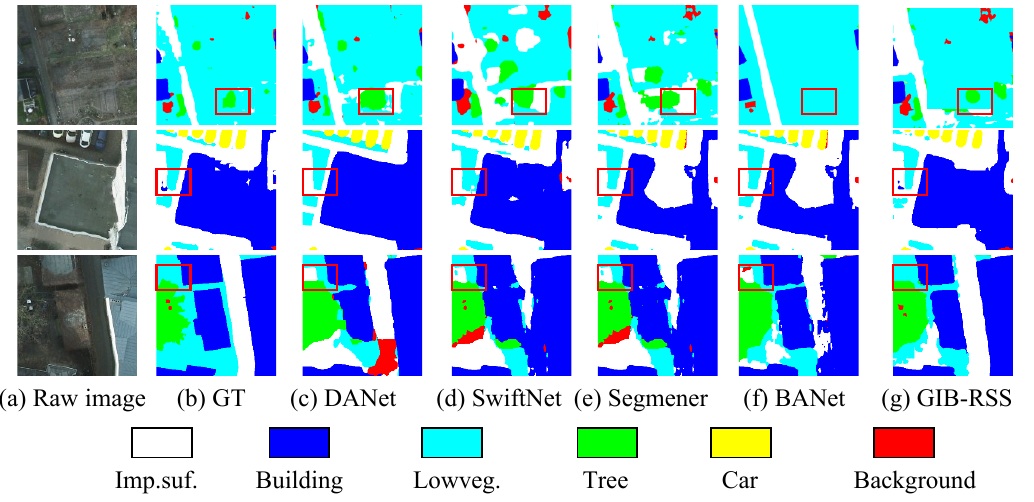}
	\caption{Visualization of the segmentation results of different models on the Postdam dataset.}
	\label{fig:postdam}
\end{figure*}

\begin{figure*}[htbp]
	\centering
	\includegraphics[width=1\linewidth]{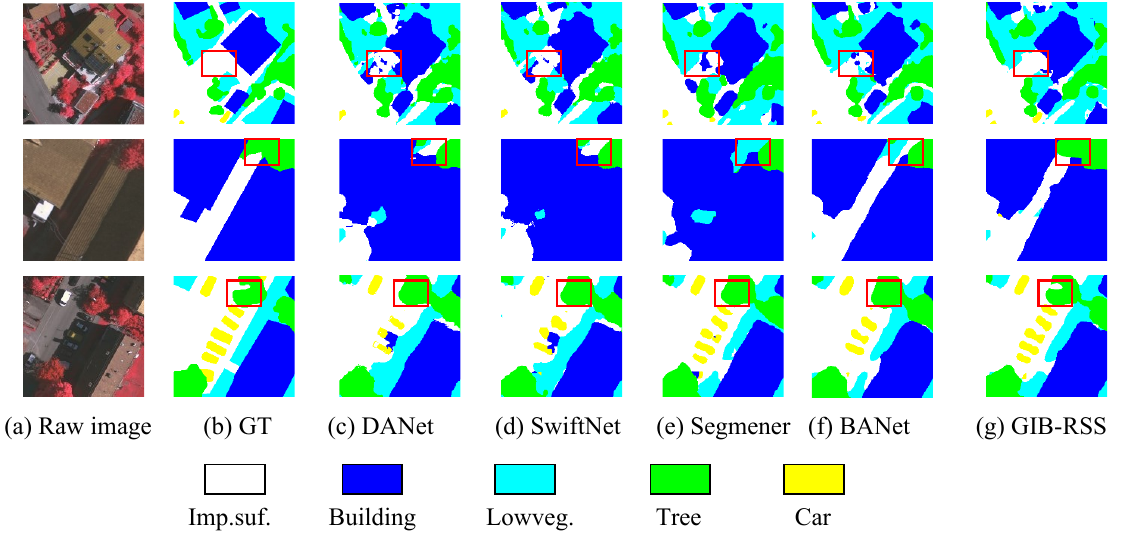}
	\caption{Visualization of the segmentation results of different models on the Vaihingen dataset.}
	\label{fig:vaihingen}
\end{figure*}

\section{Results and Discussion}
To illustrate the superiority of our proposed method GIB-RSS, we conduct experiments on four benchmark datasets (i.e., UAVid, Vaihingen, LoveDA, and Potsdam). The experimental results are shown in Tables \ref{tab:exp1}, \ref{tab:exp2}, \ref{tab:exp3}, and \ref{tab:4}. GIB-RSS outperforms the existing state-of-the-art comparison algorithms. 

Specifically, on the UAViD dataset as shown in Table \ref{tab:exp1}, GIB-RSS's mIoU value is 70.6\%, which is 3\% to 11\% higher than other models. The segmentation accuracy in other categories is also better than other comparison algorithms. For example, the IoU values of segmentation on cluster, road, tree, and vegetation have all reached SOTA, which is significantly better than existing methods. Although the IoU values on building, movingcar and human are not optimal, the difference from the best segmentation results is relatively small. Among other comparison algorithms, UNetFormer's effect is slightly lower than our algorithm, with an mIoU value of 67.8\%. We believe this is due to the fact that the architecture we designed is more suitable for segmenting irregular objects. Except for UNetFormer, the mIoU values of other comparison algorithms are significantly lower than the method GIB-RSS proposed in this paper.

On the Vaihingen dataset as shown in Table \ref{tab:exp2}, GSIB-RSS's mIoU value is 85.3\%, which is 2\% to 6\% higher than other models. OA and meanF1 values are also higher than other methods.  Specifically, the segmentation IoU value of our method GIB-RSS in four categories is significantly better than that of other comparison algorithms. It is only lower than some comparison algorithms (e.g., UNetFormer and segmenter, etc.) in the tree category. The effect of UNetFormer is second, its mIoU value is 67.8\%, which is 1.8\% lower than GIB-RSS. The segmentation effects of other comparison algorithms are significantly lower than GIB-RSS and UNetFormer, even if they use some pre-trained models with better performance.

On the Potsdam dataset as shown in \ref{tab:exp3}, GIB-RSS's mIoU value is 87.8\%, which is 1\% to 12\% higher than other models. Our algorithm GIB-RSS is significantly better than other comparison algorithms in the segmentation effects of all categories. Similarly, UNetFormer has the second best segmentation effect on the Potsdam dataset, with an mIoU value of 87.4\%. Other comparison algorithms usually use pre-trained models such as ResNet or ViT as backbone to fine-tune downstream tasks. Although the segmentation effect on the Potsdam data set is okay, it is lower than GIB-RSS.

On the LoveDA dataset as shown in \ref{tab:4}, GIB-RSS can achieve optimal segmentation results in all categories. In addition, GIB-RSS has a model parameter volume of 34.2M and an inference speed of 122.1fps, which is far superior to other comparison algorithms. Like other comparison algorithms, due to the use of large-scale pre-training models, this results in a relatively large number of model parameters and slow inference speed.

The performance improvement may be attributed to our method's ability to flexibly model irregular objects, and the introduction of the multi-head attention effectively improves the model's capture of key position information in the image. At the same time, we also introduced the information bottleneck theory to perform graph comparison learning. Unlike the previous GCL method, GIB obtains optimal graph structure representation by minimizing the mutual information between nodes. The intuition behind this is that a good augmented multi-view should be structurally heterogeneous but semantically similar. However, the existing methods are all based on CNN or Transformer architecture, and their ability to extract global position information of irregular objects is worse than GNN.

\section{Visualization of Segmentation Results}
\begin{figure*}
	\centering
	\includegraphics[width=1\linewidth]{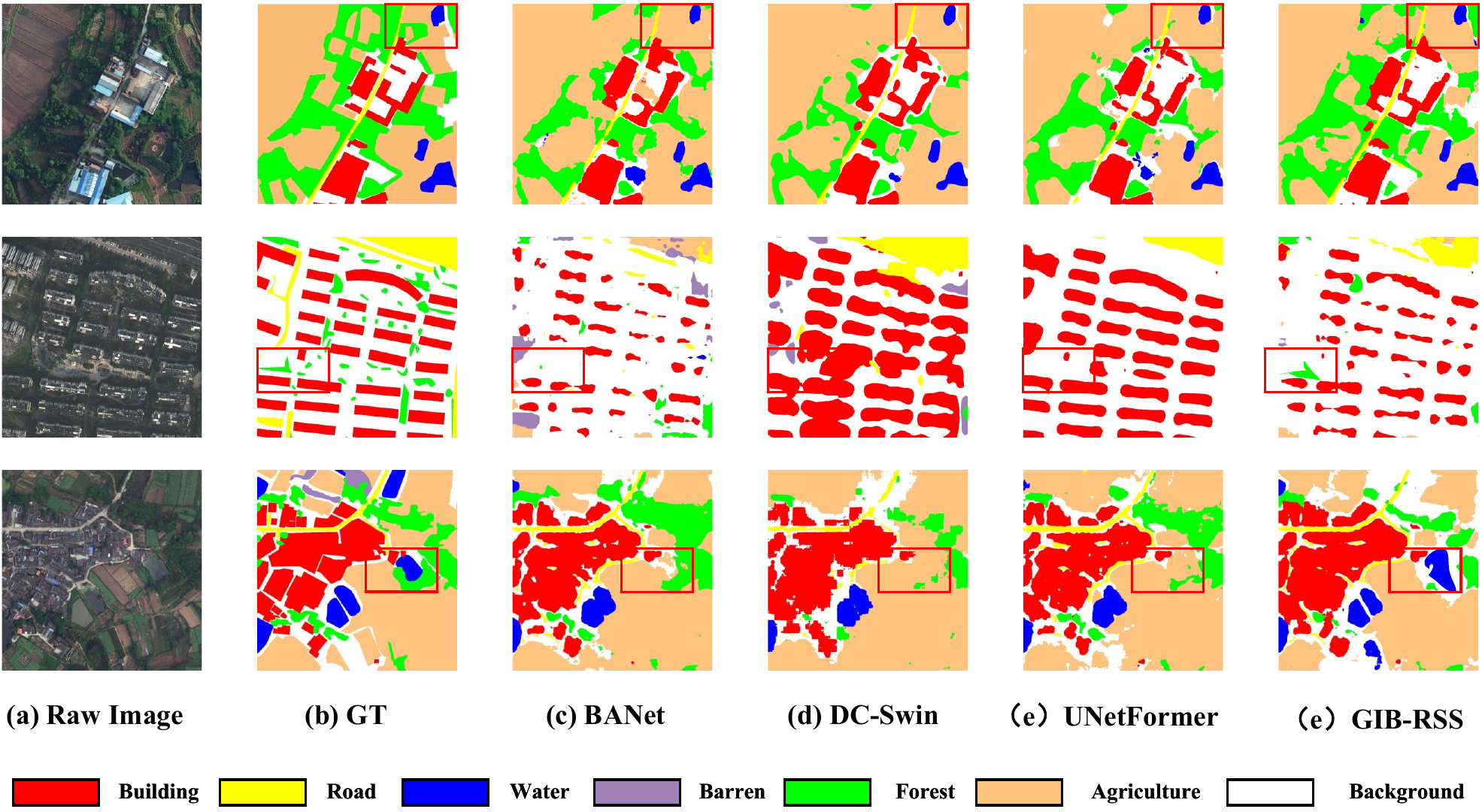}
	\caption{Visualization of the segmentation results of different models on the LoveDA dataset.}
	\label{fig:loveda}
\end{figure*}

As shown in Figs. \ref{fig:postdam}, \ref{fig:vaihingen}, and \ref{fig:loveda},  we also intuitively display the segmentation results of the model. The visualized segmentation results demonstrate the effectiveness of our designed GIB-RSS in dealing with challenging irregular objects.

Specifically, in Fig. \ref{fig:postdam}, we see that GIB-RSS can more accurately segment trees and buildings than other SOTA models, and the cases of wrong segmentation are relatively small. Other models easily misclassify lowveg. categories as background categories, and they fail to learn better for building category boundaries. In particular, in the first row of images, existing methods cannot segment the tree category well, either identifying it as background or identifying it as other categories. In the second row of pictures, existing comparison methods cannot segment some relatively small categories well, while GIB-RSS can segment small irregular objects better. In the third row of images, GIB-RSS can better segment the boundary areas of two different categories.

As shown in Fig. \ref{fig:vaihingen}, our proposed model is more clearly distinguish the difference between trees and lowveg. The experimental results show that GIB-RSS  more effectively learn the boundary information between different categories. The class boundary learning ability of other models is significantly worse than GIB-RSS. Specifically, in the first row of images, our method can better identify the background area, while other comparison methods easily misclassify the background area as a building category.  In the second row of pictures, GIB-RSS can segment the tree category relatively completely, while other methods easily identify the tree category as a background category or other categories. In the third row of images, GIB-RSS can sensitively detect the boundary areas of categories, while other methods cannot correctly segment the boundary areas of categories.

As shown in Fig. \ref{fig:loveda}, in the first row of images, existing methods cannot correctly classify the tree category, but incorrectly classify it as the agriculture category. Unlike contrastive methods, GIB-RSS can well distinguish the difference between two categories and achieve better class boundary segmentation. In the second row of pictures, since the segmented objects are relatively small, existing methods cannot perform fine-grained segmentation on them. GIB-RSS can segment small objects at fine granularity while also distinguishing differences between tree and background categories. In the third row of images, none of the existing comparison methods can segment the water category, while GIB-RSS can segment them accurately. Experimental results demonstrate the superior segmentation performance of the GIB method for irregular objects.

\section{Ablation Study}
We conduct ablation studies of our model GIB-RSS on four segmentation datasets to illustrate the effectiveness of our used modules.

\begin{table}[htbp]
	\centering
	\caption{Experimental results of different types of graph convolutional neural networks on datasets. We choose the mIoU value as our evaluation metric.}
	\label{tab:exp4}
	\renewcommand\arraystretch{1.5}
	\setlength{\tabcolsep}{9pt}{
		\begin{tabular}{ccccc}
			\toprule
			GraphConv & UAVid & Vaihingen & Potsdam & LoveDA \\ \midrule
			EdgeConv  & 69.5  &  84.4     & 86.9  &   53.6   \\
			GIN       & 68.7  &  83.6     & 86.7  &  53.1  \\
			GraphSAGE & 68.6  &  83.4     & 86.2  &  52.7   \\
			GAT       & \textbf{70.6}  & \textbf{85.3}  & \textbf{87.8}  & \textbf{54.1} \\ \bottomrule
	\end{tabular}}
\end{table}

\subsection{Type of graph convolution}
In experiments we explore the performance of three different graph convolution variants on segmentation, including EdgeConv, GIN, GraphSAGE, and GAT. As shown in Table \ref{tab:exp4}, GAT achieves the highest accuracy with mIoU values of 79.6\%, 85.3\%, 87.8\% and 54.1\% on the four datasets. The effect of EdgeConv is second, with mIoU values of 69.5\%, 84.4\%, 86.9\% and 53.6\% on the four datasets. The effect of GraSAGE is worst, with mIoU values of 68.6\%, 83.4\%, 86.2\% and 52.7\% on the four datasets. The performance improvement may be attributed to GAT's ability to capture key region information in the image.

\subsection{The effects of modules in GIB-RSS}
To illustrate that the modules (i.e., node-masking and edge-masking) proposed in this paper can better improve the performance of GNN in the field of image segmentation, we verify the effect of these modules through ablation studies. We use node mask view and edge mask view with information bottleneck criterion to improve the generalization ability of the model. From Table \ref{tab:exp5} we can see that the performance of image segmentation using graph convolution alone is not competitive. The accuracy of segmentation can be improved by introducing node-masking and edge-masking. Specifically, the model performs best when using both node mask and edge mask views, with mIoU values of 70.6\%, 85.3\%, 87.8\% and 54.1\% respectively. When only using the node mask view, the effect of the model is second, with mIoU values of 68.2\%, 82.3\%, 86.5\% and 53.8\% respectively. The model has the worst performance when the node mask and edge mask views are not used, with mIoU values of 67.5\%, 81.8\%, 86.2\% and 53.0\% respectively.

\begin{table}[htbp]
	\centering
	\caption{The effectiveness of the proposed three core modules (i.e., FFN, multi heads, and GraphConv) is verified by ablation experiments on the dataset.  We choose the mIoU value as our evaluation metric.}
	\label{tab:exp5}
	\renewcommand\arraystretch{1.5}
	\setlength{\tabcolsep}{1pt}{
		\begin{tabular}{ccccccc}
			\toprule
			GraphConv        & Node-masking       & Edge-masking              & UAVid & Vaihingen & Potsdam &  LoveDA \\ \midrule
			\CheckmarkBold & \XSolidBrush   & \XSolidBrush  &  67.5     &   81.8        &  86.2  & 53.0    \\
			\CheckmarkBold & \CheckmarkBold & \XSolidBrush   &  68.0     & 81.7          & 85.4   & 53.6    \\
			\CheckmarkBold & \XSolidBrush   & \CheckmarkBold & 68.2      &  82.3         &  86.5   &  53.8  \\
			\CheckmarkBold & \CheckmarkBold & \CheckmarkBold & \textbf{70.6}  & \textbf{85.3}  & \textbf{87.8}  & \textbf{54.1}     \\ \bottomrule
	\end{tabular}}
\end{table}

\subsection{The number of neighbors}
The number of neighbor nodes $K$ is a hyperparameter controlling information aggregation. Too few neighbor nodes will lead to low frequency of information exchange, and the global position information cannot be fully extracted, while too many neighbors will lead to over-smoothing of the model. Based on the above analysis, we adjusted the range of $K$ from 3 to 18, and the results are shown in Table \ref{tab:exp6}. When the number of neighbor nodes $K$ is 15, the segmentation effect is better. When the number of nodes $K$ is less than 15, the effect of the model increases as the number of $K$ increases, and when $K$ is greater than 15, the training effect of the model begins to show a downward trend. The above phenomenon is consistent with our analysis.

\begin{table}[htbp]
	\centering
	\caption{The influence of different number of neighbor nodes $K$ on the experimental results.  We choose the mIoU value as our evaluation metric.}
	\label{tab:exp6}
	\renewcommand\arraystretch{1.5}
	\setlength{\tabcolsep}{12pt}{
		\begin{tabular}{ccccc}
			\toprule
			$K$ & UAVid & Vaihingen & Potsdam & LoveDA  \\ \midrule
			3   & 66.7      &  82.5         &  84.4   & 52.1   \\
			6   & 67.6      &  82.9        &  85.8    &  52.7 \\
			9   & 67.8      &  83.3         & 85.9    &   52.9 \\
			12  & 68.9      &  84.2         & 86.6    &  53.6  \\
			15  & \textbf{70.6}  & \textbf{85.3}  & \textbf{87.8}  & \textbf{54.1}  \\
			18  & 69.3  &  83.9   & 87.0    &  53.6   \\ \bottomrule
	\end{tabular}}
\end{table}

\section{Conclusions}
In this paper, we regard images as graph data and introduce GNN to perform remote sensing image segmentation tasks, which can flexibly model irregular objects. To extract the global contextual location information in the image, we introduce a multi-head attention mechanism for global information extraction. Furthermore, we introduce a feed-forward network for each node to perform feature transformation on node features to encourage information diversity. In addition, in order to accelerate the convergence speed of GNN, we introduce the information bottleneck theory for graph comparison learning. We argue that a good augmented view should be structurally heterogeneous but semantically similar. Experimental results prove the superiority of our model GIB-RSS.

In the follow-up research, We hope to introduce some large-scale pre-trained models to achieve zero-shot image segmentation.

\bibliographystyle{IEEEtran}
\bibliography{refs}

\end{document}